# The State of the Art: Ontology Web-Based Languages: XML Based

Mohammad Mustafa Taye
Department of Software Engineering
Faculty of Information Technology
Philadelphia University,
Amman, Jordan

**Abstract** —— Many formal languages have been proposed to express or represent Ontologies, including RDF, RDFS, DAML+OIL and OWL. Most of these languages are based on XML syntax, but with various terminologies and expressiveness. Therefore, choosing a language for building an Ontology is the main step. The main point of choosing language to represent Ontology is based mainly on what the Ontology will represent or be used for. That language should have a range of quality support features such as ease of use, expressive power, compatibility, sharing and versioning, internationalisation. This is because different kinds of knowledge-based applications need different language features. The main objective of these languages is to add semantics to the existing information on the web. The aims of this paper is to provide a good knowledge of existing language and understanding of these languages and how could be used.

**Index Terms** — Ontology, Ontology language, RDF, RDFS, DAML+OIL , OWL

## 1 INTRODUCTION

Ontologies become a critical part in many areas, especially in Web Semantics. Consequently, a number of representational formats have been proposed to support and express them completely.

Current languages used to express Ontologies fall generally into three categories [1]: vocabularies of Ontology defined using natural language, frame-based languages used to build the structure of Ontologies based on explicit statements of class and slot, and those languages based on logic, such as Description Logics.

The main object of semantic web languages is to add semantics to the existing information on the Web. RDF/RDFS [7], OIL [14], DAML+OIL [9], and OWL [2] are modelling web languages that have already been developed to represent or express Ontologies. In general, most of these languages are based on XML [16] syntax, but they have different terminologies and expressions. Indeed, some of these languages have the ability to represent some logical relation but others do not. Because some languages have greater expressive power than others, languages chosen for representing Ontologies are based mainly on what the Ontology represents or what it will be used for. In other words, different kinds of Ontological knowledge-based applications need different language facilitators for enabling reasoning on Ontology data. These description languages provide richer constructors for forming complex class expressions and axioms.

In fact, recently most Ontology developers have used Ontology Editors, which are environments or tools used directly for editing, developing or modifying Ontologies. These tools are used for providing support for the Ontological development process, as well as for conceptualising the Ontology; they transform the conceptualisation into executable code using translators. So the output Ontology of these tools will be in one of the Web Ontology languages supported by editors Such as Protégé [18], OWL-P [21] and OilEd [22]. Alternatively Ontology reasoners are used to check the conflicts with a high degree of automation. Many such systems, including RACER [19] and FaCT [20], have been developed recently.

Returning to the main concern of this section, modelling web languages, there are in general two different types of language: Presentation languages such as HTML, designed to represent text and images to users or requesters without reference to the content, and Data Languages, intended to be processed by machines. The present research relates to the latter.

Before OWL, much research has been conducted into creating a powerful Ontology modelling language. This research stream began with the XML-based RDF and RDF/S, progressed to the Ontology Inference Layer (OIL), and continued with the creation of DAML+OIL, the result of joining the American proposal DAML-ONT5 with the European language OIL. All these languages are XML or RDF syntax based, and are consequently compatible with web standards. Indeed, RDF and OWL make searching for and reusing information both more reliable and easier, because they are considered as standards that enable the Web to be a global infrastructure for sharing documents and data equally.

As mentioned in [2], some important requirements for quality support should be taken into account when developing languages for encoding Ontologies. These include giving the user explicit written format, ease of use, expressive power, compatibility, sharing and versioning, internationalisation, formal conceptualisations of domains models, well defined syntax and semantics, efficient reasoning support, sufficient expressive power, and convenience of expression.

Syntax is one of the most important features in any language, so it should be well-defined, it is also the most significant condition required for the processing of information by machine.

The semantics of knowledge should be well defined because it represents the meaning of that knowledge. Formal semantics should be established in the domain of mathematical logic in, clearly defined way that will lead to unambiguous meaning. Well-defined semantics will lead to correct reasoning. Semantics can be considered as pre- requisites that help to support reasoning. On the other hand, reasoning will help to check and discover consistent Ontology, to verify unintended relationships between classes, and to classify individuals into classes.

This paper has detailed the most common and important languages such as RDF, RDF/S, DAML+OIL, OWL, all of which are based on XML.

Extensible Markup Language (XML) [16] is widely known in the WWW community because it is a flexible text format and was designed to describe data and to meet the challenges of large-scale e-business and electronic publishing, and it plays an important role in exchanging different type of data on the Web. In fact, it is the basis for a rapidly growing number of software development activities.

The present research, for the sake of simplicity, deals with RDF and RDFS as one description language.

## 2 RESOURCE DESCRIPTION FRAMEWORK (RDF)

Resource Description Framework (RDF) [3, 4, and 5] a language used to provide a standard for metadata about the resources on the web, is capable of representing data on and exchanging knowledge over the Web. It was developed to be understood by computers, facilitating interoperability between applications. In other words, it is a framework for using and representing metadata and describing the semantics of information about web resources in a way accessible to machines.

RDF is recommended by the World Wide Web Consortium (W3C). Uniform Resource Identifiers (URIs) are the method used by RDF to identify resources or thing. In fact, it is based built upon XML, but while that is designed for syntax, RDF is intended for semantics.

As has been mentioned, RDF is a framework for describing web resources, which is why it has become a common method of describing the properties, time, information and content of web resources, so that it can be read and understood by computer applications.

RDF can be used in several applications, one of the most important being resource discovery, used to enhance search engine capabilities. It is also used to facilitate knowledge sharing and exchange in intelligent software agents so-called Ontology, and, previously mentioned, to describe the content and content relationships available with any resource such as a page.

The RDF model has three elements: a resource (the subject), the object and the predicate. We can say that <subject> has a property <predicate> valued by <object>.

For example, a triplet could be "H.ZEDAN is the Head of the STRL Group". In an RDF graph, all triplets "nodes and arcs" should be labelled with qualified URIs. In this example, it could be said that the Subject (Resource) is the STRL Group, the predicate (property) is Head of, and the Object (literally) is H.ZEDAN.

## 2.1 RESOURCE DESCRIPTION FRAMEWORK (RDF) SCHEMA

The Resource Description Framework / Schema (RDFS) [6, 7] has been built upon the XML and RDF models and upon syntax. RDFS provides additional facilities to support evolution in both of the individual RDF vocabularies, and the core RDF Schema vocabulary.

The RDF Schema provides a machine-understandable system for defining the vocabularies needed for such applications or descriptive vocabularies. In other words, the RDF Schema is a collection of RDF resources that can be used to define or describe properties of other RDF resources which define application-specific RDF vocabularies. At the same time, RDF(S) helps developers to describe classes and properties in a specific way and to specify relationships between those properties and classes, and it allows combinations between classes, properties, or values. In other words, RDFS is used to define RDF vocabularies.

In general, RDFS is defined in a namespace informally called 'rdfs', and identified by the URI reference http://www.w3.org/2000/01/rdf-schema#. On the other hand, RDF is defined in a namespace informally called 'rdf', and identified by the URI reference http://www.w3.org/1999/02/22-rdf-syntax-ns#.

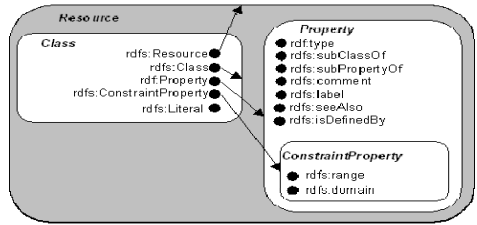

Figure 1: Elements of RDF and RDFS [7]

Figure 1 presents an excellent description of RDF /RDFS elements, demonstrating three important cores: class, property and constraint property.

In general, an RDF document contains two lists, one of descriptions and one of properties, both of which relate to one resource. Property values could be URIs, literals or others descriptions.

The rdf:RDF includes a sequence of XML elements called rdf:Description, there are "rdf:about" and "rdf:ID". In any source we need to use only one of those attributes. rdf:about is used to describe any resource; its value either an absolute or a relative URI. rdf:ID, is used to define a resource; so its value of a fragment to be added to the XML document URI.

The elements of RDF are Resource, Property, and Property Value which there are: firstly, Resource is anything that can have a URI, http://www.dmu.ac.uk/RDF, Property is for named resources such as "university", the property value is the value of a property, such as "De Montfort". For example:

```
<?xml version="1.0"?>
 <RDF>
  <Description
about="http://www.dmu.ac.uk/RDF">
    <university>De Montfort</university>
    <location>UK-Leicester</location>
  </Description>
 </RDF>
```

The statement of RDF is a combination of a resource, a property, and property value forms, as in the statement: "The University of Http://www.dmu.ac.uk/RDF is De Montfort".

The subject of the statement above is http://www.dmu.ac.uk /RDF.

The predicate is university.

The object is De Montfort.

Statement: "The location of http://www.dmu.ac.uk/RDF is UK-Leicester".

The subject of the statement above is: http://www.dmu.ac.uk/RDF.

The predicate is: location.

The object is: UK-Leicester.

## 2.2 CORE CLASSES

The elements in this class could be called fundamental concepts because they are used to describe most classes and their properties with examples. This class is defined as part of the RDF Schema vocabulary.

### 2.2.1 RDF:RESOURCE

RDF is about describing resources or any objects. Therefore, whatever is described by RDF expressions is called resources; these are always identified by URIs and are considered to be instances of the class rdfs:Resource.

### 2.2.2 RDF:PROPERTY

rdf:Property is used to represent that subset of RDF resources called properties.

### 2.2.3. RDFS:CLASS

RDF classes are the class of resources used to describe (or which could represent) anything such as cars or persons. When a schema defines a new class, the resource representing that class must have an rdf:type property whose value is the resource rdfs:Class.

```
<?xml version="1.0"?>
 <rdf:RDF
   xmlns:rdf= "http://www.w3.org/1999/02/22-rdf-syntax-ns#"
   xmlns:rdfs="http://www.w3.org/2000/01/rdf-schema#"
   xml:base= "http://www.cars.fake/cars#">
    <rdfs:Class rdf:ID="cars" />
    <rdfs:Class rdf:ID="Benz">
     <rdfs:subClassOf rdf:resource="#cars"/>
    </rdfs:Class>
 </rdf:RDF>
```

## 2.3. CORE PROPERTIES

These properties are in fact considered as an instances of the class rdf:Property. They are used to provide a mechanism for expressing relationships between classes and super-classes or between classes and their instances.

### 2.3.1. RDF:TYPE

This element is an instance of the RDF

properties used to determine that a resource is an instance of a class. In other words, this shows that a resource is a sub-set of a class, and therefore has all the features that are to be expected of a member of that class.

### 2.3.2. RDFS:SUBCLASSOF

This property is a transitive relation used to identify a relation between classes as sub/super-sets. Classes are therefore structured in a subset hierarchy represented by this property (rdfs:subClassOf) . In other words, this property is used to specify that a class C2 is a subclass of another class C1, with the logical consequence that every instance of C2 is also an instance of C1.

### 2.3.3. RDFS:SUBPROPERTYOF

This defines any property used to represent a relation between resources. This kind of property is a specialisation relation. rdfs:subPropertyOf is applied to properties to denote that one property is a  subset or specialisation of another.

### 2.3.4. RDFS:SEEALSO

The property rdfs:seeAlso used to denote the resources that might either have or  provide alternative  information about the subject resource over the internet.

### 2.3.5. RDFS:ISDEFINEDBY

The property rdfs:isDefinedBy is a sub-property of rdfs:seeAlso, and indicates the resource defining the subject resource.

### 2.4. CORE CONSTRAINTS

These are used to restrict the set of resources that may have a given property (the property's domain) and the set of valid values for a property (its range). A property may have as many values for rdfs:domain as needed, but no more than one value for rdfs:range.

### 2.4.1. RDFS:RANGE & RDFS:DOMAIN

rdfs:range is used to declare that the values of a property are instances of one or more classes.

rdfs:domain is used to declare that any resource that has a given property is an instance of one or more classes.

In the rest of this report, there are some properties used to support user-interface simple documentation these properties relate annotations within RDF schemas. This kind of concept is not essential, but could be useful in any application domain. They are defined in external schemas, but when they came into common use, they were permanently defined in the core schema.

### 2.4.2. RDFS:COMMENT

This is used to add comments to provide a human-readable description of a resource.

### 2.4.3. RDFS:LABEL

This is used to add a label name to provide a human-readable version of a resource name.

### 2.4.4. RDFS:CONTAINER

RDF Containers are could be defined as collections of resources.  The container nodes are represented by one of the three subclasses of rdfs:Container: rdf:Bag, an unordered collection; rdf:Seq, an ordered collection, and rdf:Alt used to chose between alternatives.

Although, RDF is a good basic language for many other languages, it is not very expressive and has limitations in describing resources, including descriptions of existence, cardinality, localised range and domain constraints or transitive, inverse or symmetrical properties. In general, as mentioned in [2], RDFS's expressive power is limited; on other hand, RDF/ RDFS provide modelling that concerns organisation of vocabularies in term of hierarchies: subclass and sub-property relationships, domain and range restrictions, and instances of classes. However, some features are still missing like specialised or defined properties of local scope and the specialisation of their characteristics of properties. It is impossible to separate some classes from each other. For example, we cannot say male and female are disjoint. But RDF Schema can only cater for subclass relationships - e.g. female is a subclass of person. On other hand, it is impossible to combine or create classes using Boolean relations. The expression of many restrictions is limited. The need consequently arose for a new language to supply all these deficiencies.

There are also many limitations to RDFS, among which are its inability to express equality and inequality, and its limited ability to define enumeration of property values. Regarding the latter, it cannot describe some relation among entities like union, intersection, complement, unique, symmetric, transitive, and inverse, and as far as constraints go it cannot apply cardinality and existence. Domain and range can only be specified globally. As a result, several languages such as OWL, DAML+OIL have been developed to meet these limitations.

## 3. ANNOTATED DAML+OIL ONTOLOGY MARK-UP

DARPA Agent Markup Language (DAML) + Ontology Inference Layer (OIL), DAML+OIL [8, 9] is a semantic mark-up language designed for use for Web resources. In fact, it has been built on RDF and RDF Schema, which is to say that it has an RDF/XML syntax based on the frame paradigm, and so DAML+OIL could be considered as a specific kind of RDF and this language then extended with more richer modelling primitives to cope with weaknesses in RDF /RFDS. To this end it uses URI to define the resources as RDF. DAML+OIL was actually developed to describe the structure of a domain, as most web-based languages describe structure in terms of classes and properties. DAML+OIL uses a Description Logic style model theory to formalise the meaning of a language [15].

Research first produced the Ontology Inference Layer (OIL), and a further effort produced DAML+OIL, an amalgamation of the American proposal DAML-ONT5, and the European language OIL.

## 3.1. ONTOLOGY INTERCHANGE LANGUAGES (OIL)

A semantic mark-up language for Web semantics has been built on RDF and RDF/S, this language providing modelling primitives used in frame based and Description Logic oriented Ontologies [14].

The following illustrate elements that could appear in DAML+OIL documents:

## 3.2. SETTING UP NAMESPACES

Because DAML+OIL is written in RDF, and RDF is written in XML, DAML+OIL exploits the existing Web standards XML and RDF, so a DAML+OIL document start with several namespace declarations using RDF, XML Namespace, and URIs.

```
<rdf:RDF
    xmlns:rdf ="http://www.w3.org/1999/02/22-rdf-syntax-ns#"
    xmlns:rdfs="http://www.w3.org/2000/01/rdf-schema#"
    xmlns:xsd
="http://www.w3.org/2000/10/XMLSchema#"
    xmlns:daml="http://www.w3.org/2001/10/daml+oil#"
    xmlns:dex ="http://www.w3.org/TR/2001/NOTE-daml+oil-walkthru-20011218/daml+oil-ex#"
    xmlns:exd ="http://www.w3.org/TR/2001/NOTE-daml+oil-walkthru-20011218/daml+oil-ex-dt#"
    xmlns  ="http://www.w3.org/TR/2001/NOTE-daml+oil-walkthru-20011218/daml+oil-ex#" >
</rdf:RDF>.
```

## 3.3. HOUSEKEEPING

The first declaration after the namespace that is an Ontology. This assertion is formulaic; the "about" attribute will typically be empty, indicating that the subject of this declaration is this document. For documentation purposes a few properties of this Ontology are given, such as daml:versionInfo, rdfs:comment, daml:imports

```
<daml:Ontology rdf:about="">
 <daml:versionInfo>$Id: daml+oil-ex.daml,v 1.8
2001/03/27 21:24:04 vehicle Exp $ </daml:versionInfo>
```

```
    <rdfs:comment>  An example vehicle Ontology
</rdfs:comment>
    <daml:imports
rdf:resource="http://www.w3.org/2001/10/daml+oil"/>
    <rdfs:label> vehicle </rdfs:label>
</daml:Ontology>
```

## 3.4. DEFINING CLASSES

The first step in Ontology definitions is to describe objects. It is useful to define some basic types. In DAML+OIL, classes are defined by using a daml:Class element  which is a subclass of rdfs:Class.

This is done by giving a name for the class, which is the subset of the universe which contains all objects of that type.

```
<daml:Class rdf:ID="Cars">
    <rdfs:label>Cars</rdfs:label>
    <rdfs:comment>
        This class of cars is illustrative of a number of
ontological idioms.
    </rdfs:comment>
</daml:Class>
```

This asserts that there is a class known as car. DAML+OIL divides the word into objects (which are elements of DAML classes) and data-type values. Data-type values are used to help define classes, but they are not DAML objects and cannot be included in a DAML object class. There are a number of types of car, including Mercedes- BENZ and BMW.

```
<daml:Class rdf:ID="BENZ">
    <rdfs:subClassOf rdf:resource="#Cars"/>
</daml:Class>
```

```
<daml:Class rdf:ID="BMW">
    <rdfs:subClassOf rdf:resource="#Cars"/>
    <daml:disjointWith rdf:resource="#BENZ"/>
</daml:Class>
```

This means that these two classes are disjoint (using the disjoint with tag): nothing can be both. It perfectly admissible for a class to have multiple superclasses: E-Class is a Car and Mercedes -BENZ.

```
<daml:Class rdf:ID="E-Class">
    <rdfs:subClassOf rdf:resource="#Cars"/>
    <rdfs:subClassOf rdf:resource="# Mercedes-BENZ"/>
</daml:Class>
```

## 3.5. DEFINING PROPERTIES

Properties are used to define binary relations between items. Properties in this language are of two kinds: the first, "daml:ObjectProperty", defines relations between objects,  and the second, daml:DatatypeProperty, defines relations between objects and their data-type values. It can be said that domains and ranges are global information about properties.

```
<daml:ObjectProperty rdf:ID="hasCar">
    <rdfs:domain rdf:resource="#Person"/>
    <rdfs:range rdf:resource="#Car"/>
</daml:ObjectProperty>
```

In general, properties that relate object properties to datatype values are members of DatatypeProperty. Strings, integers or decimal numbers are sometimes used when dealing with DatatypeProperty, which lead to such references as standard location XML Schema datatype decimal [8]. So for example to create an age property, non-negative integers must be mapped into XML Schema [9].

```
<daml:DatatypeProperty rdf:ID="age">
    <rdfs:comment>
        age is a DatatypeProperty whose range is
xsd:decimal.
        age is also a UniqueProperty (can only have one
age)
    </rdfs:comment>
    <rdf:type
rdf:resource="http://www.w3.org/2001/10/daml+oil#U
niqueProperty"/>
    <rdfs:range
rdf:resource="http://www.w3.org/2000/10/XMLSchema
#nonNegativeInteger"/>
</daml:DatatypeProperty>
```

## 3.6. DEFINING PROPERTY RESTRICTIONS

The property restriction is used to define an anonymous class, which contains all objects or things that satisfy the restriction. The following example defines an anonymous class of PhD student who has only a car.

```
<daml:Class rdf:ID="PhD Student">
    <rdfs:subClassOf rdf:resource="#UniverstyStudentl"/>
    <rdfs:subClassOf>
        <daml:Restriction>
            <daml:onProperty rdf:resource="#hasCar"/>
            <daml:toClass rdf:resource="#PHDStudent"/>
        </daml:Restriction>
    </rdfs:subClassOf>
```

A requirement of any web Ontology language is that statements about entities can be distributed along with different locations.

```
<daml:Class rdf:about="#PhD Student">
    <rdfs:comment>
        PhD Student has exactly two supervisors, ie:
    </rdfs:comment>
    <rdfs:subClassOf>
        <daml:Restriction daml:cardinality="2">
            <daml:onProperty
rdf:resource="#hasSupervisor"/>
        </daml:Restriction>
    </rdfs:subClassOf>
</daml:Class>
```

The cardinality property specifies a precise cardinality. But sometimes the cardinality must be restricted without being precisely specified. A person may have zero or one wife, but no more:

```
<daml:Class rdf:about="#Person">
    <rdfs:subClassOf>
        <daml:Restriction daml:maxCardinality="1">
            <daml:onProperty rdf:resource="#haswife"/>
        </daml:Restriction>
    </rdfs:subClassOf>
</daml:Class>
```

A minimal value for the cardinality of a property can be expressed thus: <Restriction daml:minCardinality="0">. A cardinality constraint could specify a maximum, minimum or precise number of values as well as enforcing the type of property those values must have.

## 3.7. NOTATIONS FOR PROPERTIES

In fact, several annotations are used in this proposal to illustrate various notations for properties:

```
<daml:UniqueProperty rdf:ID="hasMother">
    <rdfs:subPropertyOf rdf:resource="#hasParent"/>
    <rdfs:range rdf:resource="#Female"/>
</daml:UniqueProperty>
```

The (inverseOf) tag.

```
<daml:ObjectProperty rdf:ID="hasChild">
    <daml:inverseOf rdf:resource="#hasParent"/>
</daml:ObjectProperty>
The transitive tag:
<daml:TransitiveProperty rdf:ID="hasAncestor">
    <rdfs:label>hasAncestor</rdfs:label>
</daml:TransitiveProperty>
    <daml:TransitiveProperty rdf:ID="descendant"/>
```

Sometimes, "mom" is used as a synonym for "mother". The tag samePropertyAs allows this synonymy to be established:

```
<daml:ObjectProperty rdf:ID="hasMom">
    <daml:samePropertyAs rdf:resource="#hasMother"/>
</daml:ObjectProperty>
```

## 3.8. NOTATIONS FOR CLASSES

The following example represents the class "car" as being disjointed from the class "person", through use of the complementOf tag:

```
<daml:Class rdf:ID="Car">
    <rdfs:comment>no car is a person</rdfs:comment>
    <rdfs:subClassOf>
        <daml:Class>
            <daml:complementOf rdf:resource="#Person"/>
        </daml:Class>
    </rdfs:subClassOf>
</daml:Class>
```

A class disjointly united with a set of other classes can also be recognised. In this case, the class "person" can be identified with the disjoint union of the classes "man" and "woman".

```
<daml:Class rdf:about="#Person">
    <rdfs:comment>every person is a man or a
woman</rdfs:comment>
     <daml:disjointUnionOf
rdf:parseType="daml:collection">
       <daml:Class rdf:about="#Man"/>
       <daml:Class rdf:about="#Woman"/>
     </daml:disjointUnionOf>
</daml:Class>
```

The parseType="daml:collection" indicates that these sub-elements are to be treated as a unit.

```
 <daml:Class rdf:ID="ElegantWoman">
    <daml:intersectionOf
rdf:parseType="daml:collection">
       <daml:Class rdf:about="#ElegantThing"/>
       <daml:Class rdf:about="#Woman"/>
     </daml:intersectionOf>
 </daml:Class>
```

### 3.9. DEFINING INDIVIDUALS

Individual objects in a class can also be identified, for example Adam, a person of age 13 and shoesize 9.5:

```
<Person rdf:ID="Adam">
    <rdfs:label>Adam</rdfs:label>
    <rdfs:comment>Adam is a person.</rdfs:comment>
    <age><xsd:integer rdf:value="13"/></age>
    <shoesize><xsd:decimal rdf:value="9.5"/></shoesize>
</Person>
```

This datatype is used to parse the lexical representation into an actual value. A person has a property called hasSize, which is a size.

```
<daml:ObjectProperty rdf:ID="hasSize">
    <rdfs:range rdf:resource="#Size"/>
</daml:ObjectProperty>
```

Height is a class described by an explicitly enumerated set, which can be described using the oneOf element. Like disjointUnionOf, oneOf uses the RDF-extending parsetype="daml:collection".

```
<daml:Class rdf:ID="Size">
    <daml:oneOf rdf:parseType="daml:collection">
     < Size rdf:ID="small"/>
     < Size rdf:ID="medium"/>
     < Size rdf:ID="large"/>
         < Size rdf:ID="Extralarge"/>
    </daml:oneOf>
</daml:Class>
```

Finally, TallThing is the class of objects or things whose hasSize has the value tall:

```
<daml:Class rdf:ID="TallThing">
   <daml:sameClassAs>
    <daml:Restriction>
```

```
     <daml:onProperty rdf:resource="# hasSize "/>
     <daml:hasValue rdf:resource="#tall"/>
    </daml:Restriction>
  </daml:sameClassAs>
</daml:Class>
```

DAML+OIL has many limitations [17]: property constructors, it has no composition or transitive closure, in property types contain transitive and symmetrical, sets are the only collection type in this language (there are no bags or lists), there is no aggregation or comparison in data value, it has only unary and binary relations, and there are neither defaults value or nor variables.

## 4. WEB ONTOLOGY LANGUAGE (OWL)

Web Ontology Language (OWL) [2, 10, 11, 12, 13], which is a language for processing web information, became a W3C (World Wide Web Consortium) Recommendation in February / 2004. It has been built using RDF to remedy the weaknesses in RDF/S and DAML+OIL. It provides a richer integration and interoperability of data among communities and domains.

It can be said that there is a similarity between OWL and RDF, but the former is a stronger syntax with more machine interpretability and vocabulary language than RDF. Obviously, RDF is generally limited to binary ground predicates, and RDF Schema also has the limitation that it represents a subclass hierarchy and a property hierarchy, with the domain and range definitions of these properties. In other words, the language of OWL is more expressive than that of RDF and RDF Schema.

To cope with limitation in RDF, RDF/S and DAML+OIL, W3C's defined OWL. Indeed, OWL is an extension of RDFS; in other words, OWL builds on RDF and RDF Schema, and uses RDF' XML syntax; overall, OWL uses the RDF meaning of classes and properties. W3C's classify OWL into three sublanguages, each of which is intended to supply different aspects of these incompatibilities. OWL's sub-languages are OWL Full, OWL Lite, and OWL DL. What follows is a brief description of these sub-languages.

OWL documents are usually called OWL Ontologies, some elements of which are:

### 4.1. SETTING UP NAMESPACES

Because OWL is written in RDF, and RDF is written in XML, (in other words, OWL exploits the existing Web standards XML and RDF), so OWL documents start with several namespace declarations using RDF, XML Namespace, and URIs. rdf:RDF is the root element of a OWL Ontology, and also specifies a number of namespaces.

```
<rdf:RDF
  xmlns:owl ="http://www.w3.org/2002/07/owl#"
  xmlns:rdf ="http://www.w3.org/1999/02/22-rdf-
  syntax-ns#"
  xmlns:rdfs="http://www.w3.org/2000/01/rdf-
  schema#"
  xmlns:xsd ="http://www.w3.org/2001/XLMSchema#">
```

After rdf:RDF, some declarations to identify namespaces associated with this Ontology could be added. The effect of all of these namespaces is that such prefixes as owl and rdf should be understood as referring to things drawn from following namespaces, as for example http://www.w3.org/2002/07/owl#.

## 4.2. HOUSEKEEPING

After namespaces are established, any Ontology written in OWL may start with a collection of assertions for house-keeping purposes. These assertions are grouped under an owl:Ontology element which may contain comments, version statements, imports other Ontologies and labels, for example:

```
<owl:Ontology rdf:about="">
  <rdfs:comment>An example Currency Ontology
</rdfs:comment>
    <owl:versionInfo>$Id:
http://www.daml.ecs.soton.ac.uk/ont/currency.owl, v
1.0 trp 2007/06/13 10:35:25 Exp $</owl:versionInfo>
  <owl:imports rdf:resource="http://www.
kkkkkkkkk"/>
  <rdfs:label> Currency Ontology </rdfs:label>
</owl:Ontology>
```

The rdf:about attribute provides a reference for the Ontology. Where the value of the attribute is "", which is the standard case, rdfs:comment is used to provide a human-readable description of a resource. rdfs:label is used to provide a human-readable version of a resource name.

The versioning information "owl:priorVersion" which is part of the header information used to indicate earlier versions of the current Ontology. In fact, Ontologies are similar to any software that can be maintained or changed over time. There are three kinds of versioning information statement; none of them carry any formal meaning, but can be exploited by human readers:

- An owl:versionInfostatement generally contains a string giving information about the current version.

- An owl:backwardCompatibleWith statement contains a reference to another Ontology.

- An owl:incompatibleWith indicates that the Ontology containing is a later version of the Ontology referred to, but is not backward compatible with it.

## 4.3. DEFINING CLASSES

In OWL, classes are defined by using an owl:Class element that is a subclass of rdfs:Class. For example, a class "cars" is as follows:

```
<owl:Class rdf:ID="Cars">
  <rdfs:label>Cars</rdfs:label>
    <rdfs:comment>
     This class of cars is illustrative of a number of
ontological idioms.
    </rdfs:comment>
</owl:Class>
```

One of the power elements of OWL is a "owl:disjointWith", which is missing from RDFS, and is used to disjoint one class from others. "owl:equivalentClass" is another element that could be used used to establish equivalence between classes. Last but not last, there are two predefined classes, owl:Thing (which defines everything) and owl:Nothing, which is empty set.

## 4.4. DEFINING PROPERTIES

There are two kinds of properties in OWL. Object properties represent the relation between two objects, such as the relation (isOwnBy, own) used to represent the relation between car and person. Datatype properties represent the related objects with their data-type values such as person with phone, address and age.

```
<owl:ObjectProperty rdf:ID="isOwnBy">
    <owl:domain rdf:resource="#car"/>
    <owl:range rdf:resource="#person"/>
</owl:ObjectProperty>
```

In OWL it is possible to talk about Boolean combinations such as union, intersection, or complement of classes. For example it can be said that people in Jordan are Muslims or Christians.

```
<owl:Class rdf:ID="peopleAtJordan">
    <owl:unionOf rdf:parseType="Collection">
      <owl:Class rdf:about="# Muslims "/>
      <owl:Class rdf:about="# Christians "/>
    </owl:unionOf>
</owl:Class>
```

Some Boolean relations are supported by OWL; to declare more than one domain and range, the Boolean relational must be used. One example would be intersection and union among others, as well as inverse properties to inherit domain and range. Another element in OWL, owl:equivalentProperty, is used to define the equivalence of two classes and their properties, whereclasses have the same instances.

```
<owl:Class rdf:ID="Cars">
  <owl:equivalentClass rdf:resource="Automobile"/>
</owl:Class>
```

This is a strong relation used to map between two Ontologies.

## 4.5. DEFINING PROPERTY RESTRICTIONS

OWL uses some properties to add specification by using some properties' restriction to classes. These restrictions are used to define an anonymous class that will contain all things or objects that satisfy the restriction. For example, owl:allValuesFrom is used to specify the class of possible values, and owl:onProperty is used to specify the property. The next example illustrates that every car should be driven by a professional driver.

```
<owl:Class rdf:about="#everycar ">
 <rdfs:subClassOf>
  <owl:Restriction>
    <owl:onProperty rdf:resource="#isDriveBy"/>
    <owl:allValuesFrom
rdf:resource="#professionaldrivers "/>
    </owl:Restriction>
   </rdfs:subClassOf>
</owl:Class>
```

Another and similar property is owl:hasValue, which declares a specific value that the specified property must have. For example, a car should be driven by a person.

```
<owl:Class rdf:about="#car">
   <rdfs:subClassOf>
     <owl:Restriction>
      <owl:onProperty rdf:resource="#isDrivenBy"/>
      <owl:hasValue rdf:resource="#person"/>
     </owl:Restriction>
   </rdfs:subClassOf>
</owl:Class>
```

### 4.5.1. SPECIAL PROPERTIES

There are some other properties of property elements that can be defined directly: owl:TransitiveProperty defines a transitive property, such as "is older than", meaning among other things "the ancestor of", owl:SymmetricProperty defining a symmetric property such as "is sibling of", owl:FunctionalProperty, defining a property that has at most one unique value for each object such as age or National Insurance Number, and owl:InverseFunctionalProperty, which defines a property for which two different objects cannot have the same value, such as National Insurance Number.

```
<owl:ObjectProperty rdf:ID="isSiblingOf ">
   <rdf:type rdf:resource="&owl;TransitiveProperty" />
  <rdf:type rdf:resource="&owl;SymmetricProperty" />
   <rdfs:domain rdf:resource="#person" />
   <rdfs:range rdf:resource="#person" />
</owl:ObjectProperty>
```

Also,

```
<owl:ObjectProperty rdf:ID="own">
  <rdfs:range rdf:resource="#car"/>
  <rdfs:domain rdf:resource="#person"/>
    <owl:inverseOf rdf:resource="#isOwnBy"/>
</owl:ObjectProperty>
```

The rdf:parseType attribute is shorthand for an explicit syntax for building lists with <rdf:first> and <rdf:rest> tags.

```
<owl:Class rdf:ID="MuslimsAtJordan">
   <owl:intersectionOf rdf:parseType="Collection">
     < owl:Class rdf:ID="peopleAtJordan">
     <owl:complementOf>
        <owl:unionOf rdf:parseType="Collection">
          <owl:Class rdf:about="# Christians "/>
            <owl:Class rdf:about="#OtherReligion"/>
        </owl:unionOf>
      </owl:complementOf>
    </owl:intersectionOf>
</owl:Class>
```

## 4.6. ENUMERATIONS

owl:oneOf  is an enumeration element, used to define a class by listing all its elements.

```
<owl:oneOf rdf:parseType="Collection">
   <owl:Thing rdf:about="#January "/>
   <owl:Thing rdf:about="#February "/>
               :
               :
  <owl:Thing rdf:about="#Desember"/>
</owl:oneOf>
```

## 4.7. INSTANCES

These are instances of classes, and are defined in the same way as in RDF. For example, the National Insurance Number for one individual could be present thus:

```
<rdf:Description rdf:ID="9801002839">
  <rdf:type rdf:resource="#person"/>
</rdf:Description>
```

Another format could define this so: <person rdf:ID="9801002839"/>. Indeed, further details can also be added, such as:

```
<person rdf:ID="9801002839"/>
    <uni:age rdf:datatype="&xsd;integer">27<uni:age>
</person >
```

Also,

```
<carplat rdf:about="CI123TR">
    <isOwntBy rdf:resource=" 9801009493">
    <isOwntBy rdf:resource="9672006574">
</course>
```

## 4.8. THE CORE OF OWL LANGUAGE

This description language contains three expressive sublanguages. These sublanguages are OWL Lite, OWL DL, and OWL Full [1].

OWL Lite is the simplest version of OWL and provides classification hierarchy and simple constraints, being designed for easy implementation. In this sublanguage there is some restriction of OWL DL to a subset of language constructors, with some limitations such as an absence of explicit negation or union; restricted expressiveness is the disadvantage of this sublanguage.

OWL DL (short for: Description Logic) the name of this sublanguage shows that it has more Description Logic to represent the relation between objects and their properties. Indeed, it provides maximum expressiveness while preserving the completeness of computational properties. OWL Lite is a sublanguage of OWL DL.

The sublanguage OWL Full provides maximum expressiveness. OWL Lite and OWL DL are sublanguages of OWL Full.

TABLE (1)

COMPARISON BETWEEN ONTOLOGY LANGUAGES

| The expression | RDF/RDFS | DAML+OIL | OWL |
|---|---|---|---|
| Class | √ | √ | √ |
| rdf:Property | √ | √ | √ |
| rdfs:subClassOf | √ | √ | √ |
| rdfs:subPropertyOf | √ | √ | √ |
| rdfs:domain | √ | √ | √ |
| rdfs:range | √ | √ | √ |
| Individual | × | √ | √ |
| sameClassAs | × | √ | √ |
| samePropertyAs | × | √ | √ |
| sameIndividualAs | × | √ | √ |
| differentIndividualFrom | × | √ | √ |
| inverseOf | × | √ | √ |
| TransitiveProperty | × | √ | √ |
| SymmetricProper | × | √ | √ |
| FunctionalProperty | × | √ | √ |
| InverseFunctionalProperty | × | √ | √ |
| allValuesFrom | × | toClass | √ |
| someValuesFrom | × | hasClass | √ |
| minCardinality | √ | √ | √ |
| maxCardinality | √ | √ | √ |
| cardinality | √ | √ | √ |
| oneOf | √ | × | √ |
| disjointWith | × | √ | √ |
| complementOf | × | √ | √ |
| unionOf | × | √ | √ |
| intersectionOf | × | √ | √ |
| hasValue | × | √ | √ |
| imports | × | √ | √ |
| versionInfo | × | √ | √ |
| priorVersion | × | √ | √ |
| backwardCompatibleWith | × | √ | √ |
| | | √ | √ |
| incompatibleWith | × | | |

## 5. CONCLUSION

Ontology language is the basis of ontological knowledge systems, the definition of a system of knowledge representation language specification; it not only has a rich and intuitive ability to express and use it, but the body should be easily understood by the computer, processing and applications. Thus, a brief survey of state-of-the-art ontology languages which are used to express ontology over the Web is provided; all relevant terms were shown in order to provide a basic understanding of ontologies and of description logics, which are the basis of ontology languages. Therefore, choosing a language for building an Ontology is the main step. In other way, different kinds of Ontological knowledge-based applications need different language facilitators for enabling reasoning on Ontology data. These description languages provide richer constructors for forming complex class expressions and axioms.